\documentclass[twocolumn]{ceurart}

\usepackage[T1]{fontenc}
\usepackage[utf8]{inputenc} 
\usepackage{lmodern}
\usepackage{amssymb}
\usepackage{graphicx}
\usepackage[normalem]{ulem}
\usepackage{xcolor}
\usepackage{multirow}
\usepackage{comment}
\usepackage{tabularx}
\usepackage{pdfpages}
\usepackage{array}
\usepackage[section]{placeins}
\usepackage{float}
\usepackage{booktabs}
\makeatletter
\hypersetup{hypertexnames=false,hidelinks}
\makeatother

\copyrightyear{2025}
\copyrightclause{Copyright for this paper by its authors.
  Use permitted under CC BY 4.0.}
\conference{ASAIL 2025 Workshop @ ICAIL — 16 Jun 2025, Northwestern Pritzker School of Law, Chicago IL USA}

\title{What Are the Facts?\\Automated Extraction of Court-Established Facts
       from Criminal-Court Opinions}

\address[1]{Faculty of Law, Charles University, Prague, Czechia}
\address[2]{Faculty of Mathematics and Physics, Charles University, Prague, Czechia}
\address[3]{School of Computer Science, Carnegie Mellon University, Pittsburgh PA, USA}

\author[2]{Klára Bendová}[orcid=0000-0001-8002-6566]
\author[1]{Tomáš Knap}[orcid=0009-0009-2085-9182]
\author[1,2]{Jan Černý}[orcid=0009-0006-4873-8233]
\author[1]{Vojtěch Pour}[orcid=0009-0005-4075-6579]
\author[3]{Jaromir Savelka}[orcid=0000-0002-3674-5456]
\author[2]{Ivana Kvapilíková}[orcid=0000-0003-1479-3294]
\author[1]{Jakub Drápal}[orcid=0000-0001-9455-9013]
\cortext[1]{Corresponding author: Tomáš Knap, Faculty of Law,
Charles University, Prague, Czechia. 
Email: tomas.knap@prf.cuni.cz}

\begin{abstract}
Criminal justice administrative data contain only a limited amount of information about the committed offense. There is, however, an unused source of extensive information in continental European courts‘ decisions: Descriptions of criminal behaviors in verdicts by which offenders are found guilty. In this paper, we study the feasibility of extraction of these descriptions from publicly available court decisions from Slovakia. We use two different approaches for retrieval: regular expressions and large language models (LLM). Our baseline was a simple method employing regular expressions to identify typical words occurring before the beginning of the description and after. The advanced regular expressions approach further focused on sparing and its normalization (insertion of spaces between individual letters), typical for delineation of the description. The LLM approach involved prompting the Gemini Flash 2.0 model to extract the descriptions using a predefined set of instructions. Although the baseline identified descriptions in only 40.5\% of the verdicts of our test set, both methods significantly outperformed it, achieving 97\% with advanced regular expressions and 98.75\% with LLM and a combination of both 99.5\%. Evaluation by law students showed that both advanced methods matched human annotations in about 90\% of cases, compared to just 34.5\% for the baseline. LLMs fully matched human-labeled descriptions in 91.75\% of instances and a combination of advanced regular expression with LLM brings 92\% match.
\end{abstract}

\begin{document}
\begin{keywords}
  argument mining \sep court decisions \sep criminal behavior \sep NLP \sep LLM
\end{keywords}

\maketitle
\clearpairofpagestyles       
\cfoot{\pagemark}            
\KOMAoptions{headsepline=false, footsepline=false}
\pagestyle{scrheadings}   

\section{Introduction}

Empirical quantitative legal research is often hampered by insufficient detail of data, particularly in criminal law, where information about criminal behavior is frequently incomplete or insufficient. Administrative datasets typically record only the relevant section of the penal code, offering a general definition of the offense. However, they usually provide little information beyond this classification, making it difficult for researchers to discern the specific behaviors involved and to understand how variations in behavior shape state responses.

More detailed information about a criminal behavior is recorded in a textual form. Criminal verdicts in most continental European countries contain a description of the criminal behavior: An authoritative description of what behavior an offender is found guilty of and for which a sentence is imposed. These descriptions thus provide crucial and elsewhere unavailable pieces of information about the behavior. Criminal verdicts are available online in several European and non-European countries (e. g., Slovakia, Estonia, Moldova and China), while in others they are available to researchers (e. g., Finland).

In this paper, we show the difficulties in extracting these descriptions of criminal behaviors from court verdicts. After discussing related work and the data we employed, we present two different extraction methods and their specifics. Then we present outcomes of individual methods (and their combinations) and their reliability. We close by presenting how we worked with difficult cases and future use of LLMs in perfecting this task.

\section{Related Work}
Extracting various semantic and/or functional elements from court opinions has been established as a key task in legal text processing. This is because these loosely structured and sometimes noisy documents contain enormous amounts of useful knowledge that can be potentially utilized in many different applications. Prior research can be distinguished into two categories. First, the task may be defined as labeling small textual units, often sentences, according to some predefined type system (e. g., rhetorical roles, such as evidence, reasoning, conclusion). This approach has been applied to administrative decisions from the U.S.~\cite{walker2019automatic,zhong2019automatic}, multi-domain court decisions from India \cite{bhattacharya2019identification}, international arbitration decisions \cite{branting2019semi}, multi-\{domain,country\} adjudicatory decisions \cite{savelka2020cross}, or opinions of the European Court of Human Rights \cite{habernal2024mining}. Identifying a section that states an outcome of the case has also received considerable attention separately \cite{xu2020using,petrova2020extracting,grabmair2015introducing,bansal2016document}. In this work, we focus on detecting one specific sentence in each opinion---the one that authoritatively states the facts of the case.

Alternatively, the task could be to segment the text into a small number of contiguous parts typically comprising multiple paragraphs. Different variations of this task were applied to several legal domains from countries, such as Canada~\cite{farzindar2004letsum}, the Czech Republic \cite{harasta2019automatic}, France~\cite{boniolperformance}, or the U.S. \cite{savelka2018segmenting}. One approach to segmentation has focused on automatically identifying the rhetorical roles of sentences \cite{farzindar2004letsum,hachey2006extractive,moens2007summarizing}. In \cite{farzindar2004letsum} the authors employed linguistic markers to segment Canadian decisions into four units: Introduction, Context, Juridical Analysis, and Conclusion. A similar scheme was proposed in \cite{haravsta2019automatic}, including some additional types such as Dissent, Footnotes, or Party Claims. In \cite{wyner2010approaches}, the authors identify typical language structures that are used in various types of Premises or Conclusions. These are then expressed in the form of Context Free Grammar for parsing legal arguments. In \cite{saravanan2006improving,saravanan2010identification}, conditional random fields (CRF) were applied to segment legal documents into seven labeled components with each label representing a corresponding rhetorical role.

\section{Data}

Slovak courts' verdicts are in many respects typical of the continental Germanic legal culture. They are clearly divided into two parts: Dispositive part that presents the decisions that were taken and reasoning. Dispositive part (i) identifies the case, the court, the verdict, and the offender, (ii) announces whether the offender was found guilty or not, (iii) describes criminal behavior for which the offender was found guilty (or innocent), (iv) subsumes the behavior within a legal definition of an offense and (v) pronounces the consequences (sentences and reparation of damages). An example of this part is in Appendix B. Some verdicts further contain reasoning---an explanation of why the offender was found guilty, why a specific sentence was imposed and a description of the proceedings. Reasoning is often infrequent and of low quality \cite{drapal2024sentence}.

Slovakia allows anyone to \href{https://web.archive.org/web/20230205033430/https://obcan.justice.sk/opendata}{download .json files with all court decisions} and to use an \href{https://obcan.justice.sk/pilot/api/ress-isu-service/swagger-ui/index.html}{API to search for specific decisions}. While such convenient accessibility is very beneficial, researchers should worry about possible flaws of the published data, especially missing data and possible resulting limited representativeness. If an administrative dataset containing secondary data about court decisions made in a country is available, it helps researchers to identify both how many court verdicts are missing and whether there is a pattern within missing values. Slovakia has an administrative dataset of high quality containing secondary data describing all criminal court verdicts from 2018 to 2022 (henceforth only "administrative dataset") \cite{terkanivcrevizia}. There were 126,795 verdicts decided during this period according to this dataset.

We first downloaded all court verdicts and we linked the verdicts with the administrative dataset using a unique court docket number and court name. We were able to link 77.64\% of cases. Then we employed the court docket numbers from the administrative dataset to retrieve the missing verdicts via API, which allowed us to link an additional 12.42\% of cases. Overall, we were not able to link 9.94\% of cases, which should exist, yet they are (i) not present in the .json or API, (ii) present in .json but matching fails due to a corrupt .json file or (iii) present in API but matching fails due to repeating requests and website limitations. Successful matching to an extent depended upon which court decided the case. While for 40 out of 54 district courts the success rate was higher than 90\%, other courts had a success rate higher than 60\% with the exception of one, which had a success rate of 12\%. These differences present the need to work with administrative datasets to ensure full-text court verdicts are representative and to determine its limitation. This provides us with 112,864 court verdicts within which we attempted to identify descriptions of criminal behavior.

\begin{table}[htbp]
  \centering\scriptsize
  \caption{Coverage of Slovak criminal-court verdicts (2018–2022)}
  \label{tab:coverage}
  \begin{tabular}{lrr}
    \hline
    \textbf{Download / linkage step} & \textbf{n cases} & \textbf{\% of 126\,795} \\
    \hline
    JSON dump (direct match)              &  \phantom{1}98,500 & 77.64\,\% \\
    API re-download (matched via docket number)  &  \phantom{1}15,742 & 12.42\,\% \\
    \hline
    \textit{Linked total}                 & 112\,864 & 88.06\,\% \\
    Not linked (missing / corrupt / API errors) & 13\,931 & 11.0\,\% \\
    \hline
  \end{tabular}
\end{table}

We chose two test sets of verdicts to annotate the data. The first, general stratified sample contains 400 judgments where we controlled for years and representation of different courts during sampling. These were then annotated by trained law students (2 groups by 200 statements) with each statement being annotated by two persons (agreement of 97\% in both groups, disagreements resolved by one of the authors). This dataset is intended to evaluate the overall performance of the individual methods. Since we expected high performance on the general dataset, we also created an additional set of 200 judgments to evaluate more challenging cases. This supplementary set includes a mix of judgments in which the sparing extraction yielded only a single candidate expression - indicating a higher likelihood that the judgment lacks a factual sentence - and judgments in which the rule-based extraction method failed to identify any relevant sentence at all. The data were annotated under the same conditions (95.5\% agreement; disagreements were resolved by one of the authors).

\begin{table}[!htbp]
  \centering
  \footnotesize
  \setlength{\tabcolsep}{3pt}     
  \renewcommand{\arraystretch}{0.95} 
  \caption{Annotated evaluation sets}
  \label{tab:testsets}
  \begin{tabularx}{\columnwidth}{@{}l c >{\raggedright\arraybackslash}X c >{\raggedright\arraybackslash}X@{}}
    \toprule
    Set & n cases & Sampling & Agree. & Purpose \\
    \midrule
    General stratified & 400 & year × court strata & 97\% & overall performance \\
    Challenging cases  & 200 & sparing = 1 $\vee$ regex fail & 95.5\% & stress-test edge cases \\
    \bottomrule
  \end{tabularx}
\end{table}

\section{Experimental Design}

\subsection{Structure of court decision}
\label{subsec:structure_decision}
Our analysis leveraged two fundamental formal features observable across all judgments: A consistent structure (described above) and the systematic use of a specific typographic convention throughout the dataset, specifically sparing. The letters of a word are spaced out (e. g., L I K E T H I S). Sparing is generally applied to single words or short phrases and is typically placed between paragraphs. Its function is to introduce a new section of the judgment. As a result, sparing tends to appear with a limited set of expressions used across decisions as we can see in Appendix B. A key advantage of sparing is its robustness during format conversion: even when judgments are somewhat messily transformed from PDF to .json, sparing is usually preserved---unlike paragraph-ending characters, which often suffer from noisy encoding. However, sparing has a domain-specific nature; while common in legal documents, it is rarely encountered in other textual domains.

\subsection{Baseline}


As we describe in Section ~\ref{subsec:structure_decision}, both the overall structure and the sparing expression between paragraphs appear to be consistent at first glance. This observation motivated us to extract factual statements using regular expressions because of their simplicity, accuracy, and ease of use. In the baseline approach, we chose not to introduce any complexity into the regex patterns. We manually listed starting and ending phrases as fixed patterns (examples of phrases are in Table \ref{tab:Starting_Ending_expression}) and used them for extraction. The baseline achieves a fact sentence extraction success rate of 31.69\%.

\begin{table}[!htbp]
\caption{Typical starting and ending expressions}
\label{tab:Starting_Ending_expression}
\centering
\scriptsize
\begin{tabular}{@{}l l@{}}
\hline
\textbf{Starting Expression} & \textbf{Ending Expression} \\
\hline
is found guilty that         & therefore/thus \\
is found guilty              &                 \\
they are guilty that         &                 \\
is acknowledged as guilty that &               \\
is acknowledged guilty that  &                 \\
\hline
\end{tabular}
\end{table}

    \setlength{\parskip}{0pt}

\subsection{Advanced regular expressions}

Building on observations on simple regular expressions, we developed more flexible patterns to eliminate variations in the input text. These flexible patterns focused on preventing unwanted irregularities—such as extra spaces, unexpected line breaks, and other conversion issues. Adding optional whitespace matching between characters in key phrases significantly improved resiliency to formatting inconsistencies.\footnote{ \scriptsize \url{https://github.com/vanickovak/Factual-sentence-extraction.git}} This improved both the success rate and the quality of the extracted fact sentences as shown in Table \ref{tab:method_performance}.

The regular expressions themselves were automatically extracted directly from the court verdicts. We focused only on sparing expressions from the rulings, preserving their order in the document. We then grouped expressions by their relative position and annotated those that function as openers or closers of fact sentences. This yielded a set of starting ($N = 40$) and ending ($N = 2$) points to identify fact sentences.

\begin{table}[htbp]
  \centering\scriptsize
  \renewcommand{\arraystretch}{0.95} 
  \caption{Comparison of simple regex and automatic approach extraction}
  \label{tab:regex_comparison}
  \begin{tabular}{lrrr}
    \toprule
    \textbf{Method} & \textbf{Extracted} & \textbf{Extracted (\%)} & \textbf{Total} \\
    \midrule
    Simple Regex      & 35,767  & 31.69\,\% & 112,864 \\
    Automatic approach & 109,292 & 96.80\,\% & 112,864 \\
    \bottomrule
  \end{tabular}
\end{table}

\subsection{LLMs: Standalone and building on previous methods}
During the initial phases of our research, relying solely on a rule-based approach for extracting factual statements proved insufficient. We have made a natural progression to using more sophisticated approaches. The main challenge with defining a narrow scope of rules, especially in a language where the individual words are typically more varied, is that one can hardly create a complete list of all the possible words and their combinations. LLMs are an ideal solution for extending a narrow list of rules to a dataset that presents uncertainties about the exact wording in each individual case.

Utilizing LLMs involves two main aspects: selecting an appropriate model and providing it with the necessary context, a process known as prompting. Our model selection was guided by two primary criteria: its established suitability for relevant applications such as parsing and extracting from large texts, and overall cost. Based on this evaluation, Gemini Flash 2.0\footnote{ \scriptsize \url{https://blog.google/technology/google-deepmind/google-gemini-ai-update-december-2024}} was identified as the most suitable option. Our temperature was 0.0, which helps reduce hallucinations in extracting factual sentences from judgments and supports better replicability of our research. Our initial experiments revealed that generic descriptions of factual statements were insufficient for reliable extraction using LLMs. Achieving success required a carefully constructed prompt specifically designed for this task. We developed a specialized prompt that combines concrete examples of factual statements, a clear definition of the expected output, and, importantly, explicit indicators of where these statements typically begin and end (Appendix A). 

The critical breakthrough in our approach occurred when we followed phrases used by previous methods. We incorporated specific textual markers commonly found at the beginning or end of factual statements  as shown in Table~\ref{tab:Starting_Ending_expression}.
This structural guidance transformed the task from pure content understanding to a more focused pattern recognition and extraction challenge. The model was directed to identify text between these markers and extract only the factual components while omitting legal evaluations. The results were returned in .json format for further processing.

A single Slovak judgment contains on average 4\,083 characters
(approx. 1,020 tokens), to which we prepend a 10\,497-character prompt
(approx. 2,624 tokens).  The Gemini Flash 2.0 output averages
1\,257 characters (approx. 314 tokens).
At Google’s July 2025 pricing
(USD 0.10 / M input tokens, 0.40 / M output tokens),
this corresponds to 0.00049 USD per judgment
(approx. 0.49 USD for 1,000 decisions).

Despite recent improvements in LLM accuracy, we observed occasional instances of text hallucination, particularly with special characters or uncommon legal terminology. To address this issue, we implemented a post-processing step using a function that aligns the model's output with the original text. This verification method effectively eliminated hallucinations, ensuring fidelity to the source documents.

\subsection{Combination Approach}
Our experimental design includes a combined methodology that integrates the strengths of both advanced regular expressions and LLM approaches. This hybrid pipeline operates sequentially:

1) Advanced regular expressions first attempt to extract factual statements from the court decisions.
2) For cases where the regular expression approach fails to identify any factual sentence, we apply the LLM-based extraction.

This combination strategy maximizes efficiency while addressing the limitations of each individual method. The regular expressions provide fast processing for standard document structures, while the LLM handles more complex linguistic variations and non-standard document formats.

\section{Results}
To empirically assess our approach, we tested the ability of the method to extract specific factual sentences against a set of 400 reference statements.

First, we assess the ability of the method to identify a factual statement in the court decision. While the baseline found a description of criminal behavior in only 40.5\% of verdicts, both advanced regular expressions and LLMs performed significantly better, achieving 97\% and 98.75\% success rates, respectively. If we combine both approaches, specifically by employing LLMs only on cases where advanced regular expressions fail to extract any factual sentence, we achieve an improved overall accuracy of 99.5\%. The reason why the LLM did not achieve the extraction of 100\% factual statements is that five cases could not be processed due to exceeding the model token limits or triggering content safety flags within the API, likely related to sensitive topics such as the narcotics mentioned in the cases. Feeding the texts to the model in pieces and prompt engineering to bypass the safety mechanism would likely increase the performance further.

\begin{table}[htbp]
\caption{Method performance on test data}
\centering
\scriptsize
\begin{tabular}{lcc}
\hline
\textbf{Method} & \textbf{Extraction} & \textbf{Quality} \\
& \textbf{Rate} & \textbf{$\geq$95\%} \\
\hline
Baseline & 40.5\% & 34.5\% \\
Advanced Regex & 97.0\% & 89.5\% \\
LLM & 98.75\% & 91.75\% \\
Combined & 99.5\% & 92.0\% \\
\hline
\end{tabular}
\label{tab:method_performance}
\end{table}

We further focused on the quality of the extracted sentences. Extraction quality was measured at the character level, ignoring diacritics. The results are displayed in Table~\ref{tab:method_performance} by the quality of a match. When evaluating based on exact match, the performance of the LLM is undeniable. However, if we allow for minor character-level variations of up to 5\%, the results of advanced regex and LLMs are surprisingly comparable—89.5\% for the advanced regex and 91.75\% for the LLM. In contrast, the baseline approach achieved this level of approximate matching in only 34.5\% of cases. By combining both approaches—specifically applying the LLM only to cases where advanced regex fails to extract accurate factual sentences—we achieve 92\% accuracy on the test data.

\begin{table}[htbp]
\caption{Detailed extraction quality breakdown}
\centering
\scriptsize
\begin{tabular}{lccc}
\hline
\textbf{Method} & \textbf{Perfect} & \textbf{Good} & \textbf{Failed} \\
& \textbf{100\%} & \textbf{80-95\%} & \textbf{<50\%} \\
\hline
Baseline & 2.0\% & 0.8\% & 4.5\% \\
Adv. Regex & 9.0\% & 2.8\% & 2.8\% \\
LLM & 91.8\% & 5.3\% & 1.3\% \\
Combined & 11.5\% & 2.8\% & 2.8\% \\
\hline
\end{tabular}
\label{tab:quality_breakdown}
\end{table}

We then focused on problematic cases, i.e., verdicts where no factual statement was extracted or where the extracted statement shared less than 50\% character overlap with the manually annotated version. Such cases typically result from non-standard typographic formatting of the verdict or the use of less frequent terms in sparing expressions.
In some instances, the extraction failed because the factual statement was entirely missing from the verdict, which is usually due to their procedural nature (e. g., acquittal judgments).
We prepared a dataset of 200 such verdicts exhibiting these deviations and had it annotated by two annotators under the same conditions as the main test dataset. The same LLM prompt was then applied to this data.

\begin{table}[!htbp]
\caption{LLM performance on difficult cases ($N=200$)}
\label{tab:difficult_cases}
\centering
\scriptsize
\begin{tabular}{@{}l r@{}}
\hline
\textbf{Outcome} & \textbf{Share} \\
\hline
Perfect match (100\%)        & 81.5\,\% \\
High quality (95--100\%)     & 2.5\,\%  \\
Good quality (80--95\%)      & 3.5\,\%  \\
Fair quality (50--80\%)      & 0.5\,\%  \\
Failed extraction            & 12.0\,\% \\
\hline
\textbf{Total high quality ($\geq$95\%)} & \textbf{84.0\,\%} \\
\hline
\end{tabular}
\end{table}

As shown in Table \ref{tab:difficult_cases}, the LLM was able to perform high-quality factual statement extraction in 84\% of cases within this challenging dataset, and failed to identify a factual statement in 12\% of cases, resulting in 0\% similarity. Investigation confirmed that in nearly all these instances, the LLM correctly followed its strict prompt instructions. The failures occurred because the factual sentences in this challenging subset used grammatical structures or formats not anticipated by the specific rules in the prompt. The LLM, therefore, acted as instructed by reporting no match, rather than misinterpreting the content. This highlights the need to update the prompt, generalizing the description of a factual sentence based on the wider variety of formats observed.

To analyze hallucinations, we compared 100 LLM-generated factual statements with the
corresponding spans in the source judgments.
Character-level similarity averaged 0.99; only one case
(1 \%) fell below 90 \%.

To avoid leaking hallucinated text into downstream data, we
post-process every model output: we located the predicted span in the
source document and replaced the generated string with that exact
substring.  The factual sentences stored in our dataset are therefore
\emph{verbatim} excerpts from the original judgments; hallucinations
show up only as cases where no sufficiently similar span can be
aligned.

\begin{table}[htbp]
  \centering\scriptsize
  \caption{Hallucination breakdown $(N\!=\!100)$}
  \label{tab:hallucination_breakdown}
  \begin{tabular}{lcc}
    \hline
    Similarity band & Cases & Share \\
    \hline
    100 \% & 26 & 26.0 \%\\
    95–99 \% & 71 & 71.0 \%\\
    90–94 \% & 2 & 2.0 \%\\
    80–89 \% & 1 & 1.0 \%\\
    $<$80 \% & 0 & 0.0 \%\\
    \hline
  \end{tabular}
\end{table}

\section{Discussion: Integrating all Methods}

We did not train a supervised model due to the high cost of large-scale annotations. Our
strategy was to first explore few-shot prompting of an LLM, measure
its performance against manually annotated data, and consider
supervised training only if the LLM fell short. A small experiment with training a model on the annotated test set could be interesting, but it would have left us without a clean held-out set for evaluation, making it difficult to fairly compare the approaches. 

Our goal was to develop a scalable solution that could be applied across a wide range of court decisions without the need for extensive labeled training data. To that end, we combined regular expressions with few-shot prompting of an LLM. This approach allows us to inject a small number of annotated examples directly into the prompt, guiding the model’s output in a flexible and easily adjustable way. By avoiding model training and relying instead on prompt-level supervision and rule-based heuristics, we were able to build a system that is usable in low-resource legal settings and avoids the high cost of large-scale manual annotations.

The LLM-based approach introduces computational demands that must be balanced against extraction quality. While more resource-intensive than simpler methods, the accuracy benefits justify its application, particularly for complex or non-standard documents. For large-scale processing of thousands of verdicts, we recommend a staged approach where computationally expensive LLM processing is reserved for documents where simpler methods yield low-confidence results.

It is worth noting that a recurring pattern observed in cases with imperfect matches was the LLMs' occasional difficulty in precisely delineating the boundaries of the target factual statement within the broader text. Specifically, mismatches often arose not from extracting incorrect information; rather, it was from the model including extraneous text immediately following the intended end of the factual statement, or in some other cases, the LLM has finished the factual sentence before where it was instructed.

These detailed results affirm the LLMs' precision, particularly its high success rate, but also highlight challenges with edge cases and computational intensity, which requires careful consideration for scalability. Future enhancements include employing models with larger context windows to handle longer texts without truncation, and refining prompts for greater task specificity to further improve accuracy and relevance, particularly for the less successful cases.

Finally, transferring the methods to other jurisdictions also remains a challenge. Each new legal system
would require its own regex inventory reflecting local idioms and
citation habits. The LLM approach proved more robust in our pilots, but our sample is too narrow to draw general conclusions about
out-of-domain performance without retraining and expert validation.
We therefore refrain from reporting cross-jurisdiction metrics and
leave systematic evaluation on foreign corpora for future work.

\section{Conclusion}

In this paper, we presented a method for collecting published court decisions and extracting factual sentences---coherent descriptions of criminal conduct---with strong potential for further research. In the extraction process, we identified the consistent structure of court decisions and a legal typographic convention known as ``sparing'' as key features. We leveraged these observations in three extraction approaches: an automated search for start- and end-markers of factual sentences (baseline), an advanced regular expression-based script, and, surprisingly effectively, extraction using a LLM. The baseline approach lacked the complexity to successfully extract the factual sentences from verdicts. In contrast, the advanced regular expressions identified descriptions in 97\% of verdicts and Gemini Flash 2.0 extracted them in 98.75\% of the test data. Combination of both methods extracted descriptions in 99.5\% of cases. Manual annotation revealed that 91.75\% of descriptions retrieved by the LLM and 89.5\% of those retrieved by regular expressions match the descriptions identified by human annotators. The combination of advanced regular expression and LLM achieved 92\% accuracy. In future work, we aim to expand the dataset with court decisions from additional countries, providing empirical data for comparative research in criminal law.

\begin{acknowledgments} This study was funded by the Czech Grant Foundation (grant number 25-16848M entitled "Just Sentences: Analyzing and Enhancing Proportionality and Consistency Using Typical Crimes"). The authors have no competing interests.
\end{acknowledgments}
\newpage 


%
%
%
\bibliographystyle{plainnat}
\clearpage 

\appendix


\clearpage
\appendix
\onecolumn            
\includepdf[
  pages=1,
  pagecommand={%
    \thispagestyle{plain}%
    \section*{Appendix A: Prompt}\label{app:a}%
  }
]{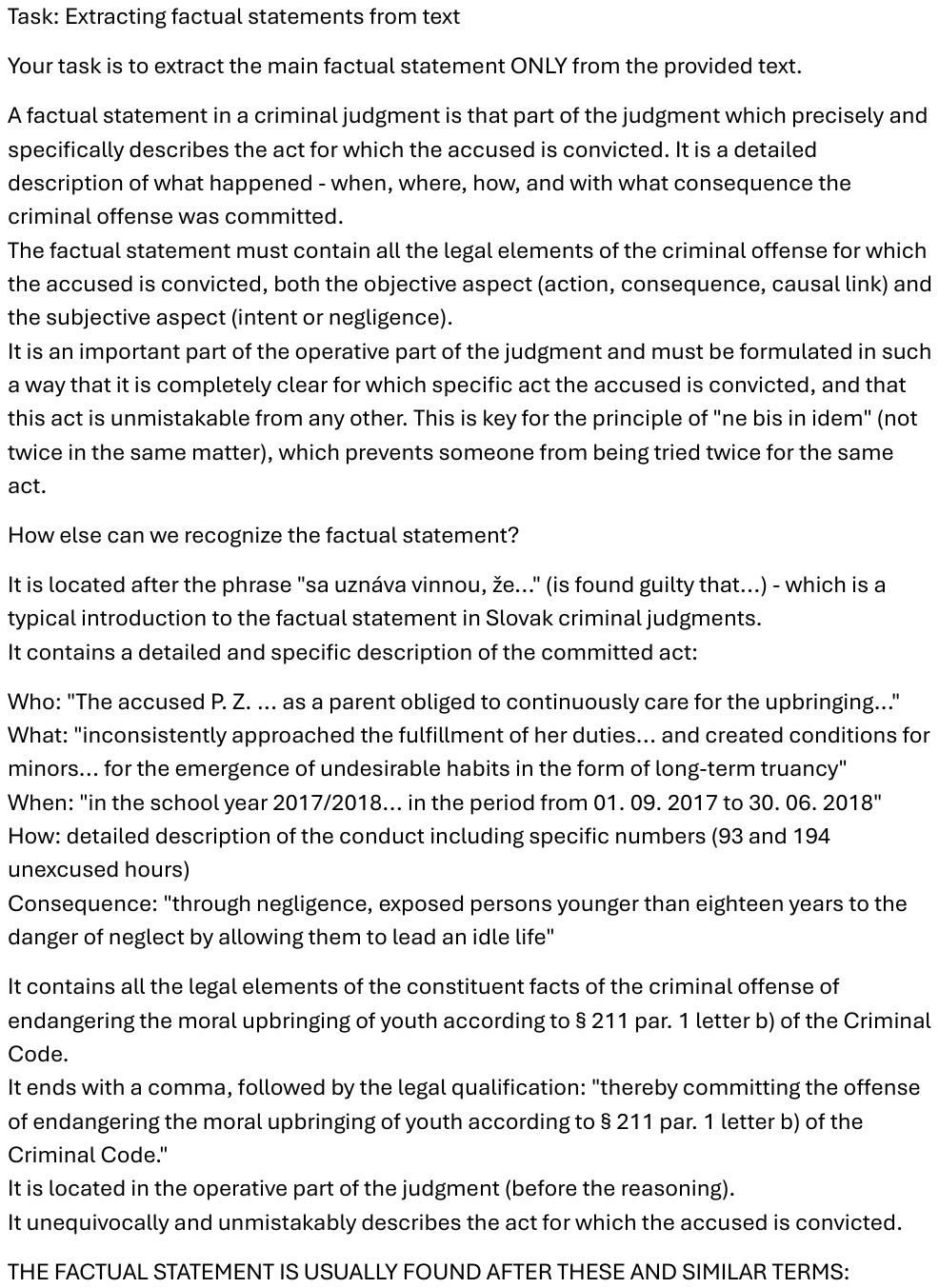}

\includepdf[
  pages=2-,
  pagecommand={\thispagestyle{plain}}
]{prompt.pdf}

\clearpage
\thispagestyle{plain}
\section*{Appendix B: Example Slovak Criminal Verdict }
\label{tab:app_b}
\begin{center}
\includegraphics[
  width=0.9\paperwidth,      
  keepaspectratio
]{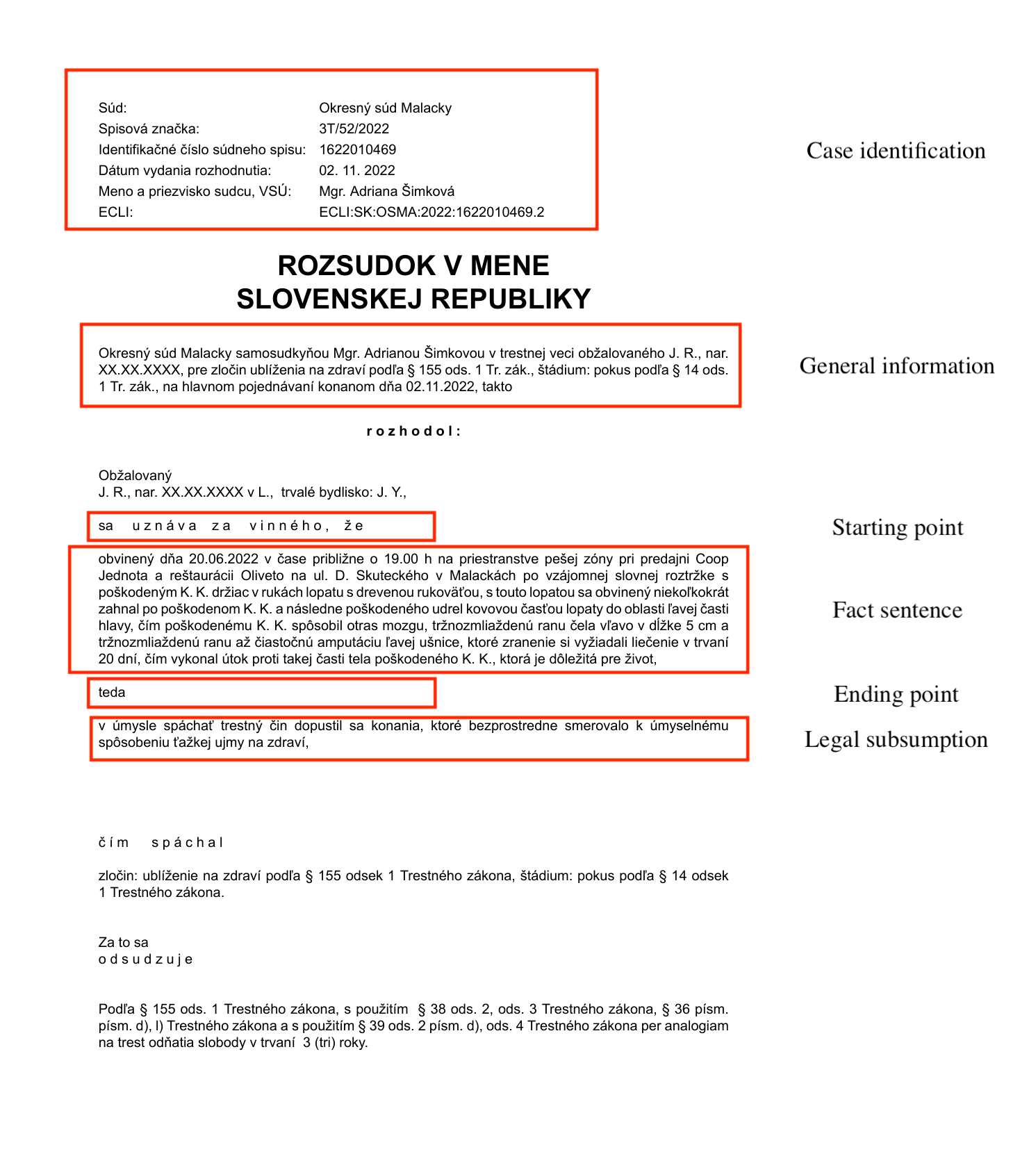}
\end{center}



\end{document}